\documentclass[]{interact}

\usepackage{epstopdf}% To incorporate .eps i llustrations using PDFLaTeX, etc.
\usepackage[caption=false]{subfig}% Support for small, `sub' figures and tables
\usepackage[ruled,vlined]{algorithm2e}
\usepackage{multirow}
\usepackage{pgfplots}
\usepackage[numbers,sort&compress]{natbib}% Citation support using natbib.sty
\bibpunct[, ]{[}{]}{,}{n}{,}{,}% Citation support using natbib.sty
\makeatletter% @ becomes a letter
\def\NAT@def@citea{\def\@citea{\NAT@separator}}% Suppress spaces between citations using natbib.sty
\makeatother% @ becomes a symbol again

\theoremstyle{plain}% Theorem-like structures provided by amsthm.sty

\theoremstyle{definition}

\theoremstyle{remark}

\begin{document}

%\articletype{ARTICLE TEMPLATE}% Specify the article type or omit as appropriate

\title{\LARGE \bf
Auxiliary Heuristics for Frontier Based Planners}

\author{
\name{Arsh Tangri\textsuperscript{a}, Dhruv Joshi\textsuperscript{b} and 
Ashalatha Nayak\textsuperscript{c}\thanks{CONTACT Arsh Tangri. Email: arsh.tangri2@gmail.com}}
\affil{\textsuperscript{a}Department of ECE, Manipal Institute of Technology,Manipal,Karnataka,India; \textsuperscript{b}Department of CSE, Manipal Institute of Technology,Manipal,Karnataka,India;
\textsuperscript{c}Department of CSE, Manipal Institute of Technology,Manipal,Karnataka,India.}}

\maketitle

\begin{abstract}
Autonomous exploration of unknown environments is a vital function for robots and has applications in a wide variety of scenarios. Our focus primarily lies in its application for the task of efficient coverage of unknown environments. Various methods have been proposed for this task and frontier based methods are an efficient category in this class of methods. Efficiency is of utmost importance in exploration and heuristics play a critical role in guiding our search. In this work we demonstrate the ability of heuristics that are learnt by imitating clairvoyant oracles. These learnt heuristics can be used to predict the expected future return from selected states without building search trees, which are inefficient and limited by on-board compute. We also propose an additional filter-based heuristic which results in an enhancement in the performance of the frontier-based planner with respect to certain tasks such as coverage planning.
\end{abstract}


\section{Introduction}

The primary goal of autonomous exploration is to build an accurate map of the environment in a finite amount of time and with minimal computational overhead. Algorithms for autonomous exploration play a vital role in the efficient coverage of unknown environments and allow autonomous machines to accurately navigate unknown terrains. The planner we study in this paper falls under a class of planners called frontier-based planners.

Frontier-based planners are planners that solve the task of autonomous exploration efficiently by selecting and travelling to the frontier point that generates the maximum reward at each step. The frontier is the boundary between the explored and the unexplored region and frontier points are locations along this boundary. There are several variations to these algorithms and changes include variations in the method of generating frontier points, the path planner used to generate the path to a selected frontier point and the heuristic function used to evaluate a frontier point.

In this paper we focus on the heuristic function and demonstrate that careful selection of this function results in consistent improvement in the performance of the planner. The heuristic plays a key role in guiding the agent and variations in this can drastically alter the path traversed by the agent and its performance. A typical heuristic used is a linear combination of the information gained while travelling from the current location to a selected frontier point and the cost of this traversal. Such heuristic lacks information about the future rewards, which can be a useful differentiator for coverage planning tasks where the future gain in information plays an important role. One of the simplest methods for evaluating future reward is to build a search tree of possible future states and to grow this tree as we look further ahead. This can be extremely inefficient as the tree becomes extremely large and hard to evaluate as the number of frontier points increases and as we look further into the future. To solve this problem, we propose a network that learns to imitate an oracle that predicts the future reward. The oracle is computed using the search tree method and it is used to generate a dataset of states and their corresponding future returns. The network is then trained on this dataset to imitate the oracle.

The heuristic so far prioritizes gain in information and although it works well for exploration tasks, it tends to leave small clusters of unexplored regions that are surrounded by explored regions. This happens because traversal of those regions does not generate high reward according to our current heuristic. This makes it slightly inefficient for mapping and coverage planning, where traversing the entire map is the primary goal. To solve this issue, we propose an additional heuristic that provides incentive for the agent to cover these unexplored clusters thus improving the planner’s performance for coverage planning tasks. The heuristic comprises of a set of filters that get activated when applied on regions that have unexplored grid cells surrounded by several explored grid cells. This allows the heuristic function to prioritize these regions during traversal. Our contributions are as follows:
\begin{itemize}
    \item We augment the heuristic function of frontier-based planners by using neural networks to predict the expected future reward for a given frontier point.  
    \item We further improve the performance of the planner by using filter based heuristics to incentivize vital regions that are overlooked by these planners.  
\end{itemize} .

\section{Related Work}

In existing research, frontier-based methods vary in the method of generation of frontier points, the path planner and the heuristic function. For the generation of frontier points, some methods are based on edge detection while some use region extraction that require processing the entire map. Some methods also make use of the latest scans or updated cells, for the same. Keidar and Kaminka \cite{article2} and Senarathne et al \cite{article3} propose a method that only consider the latest scans or the updated grid cells. Umari and Mukhopadhyay \cite{article4} make use of the Rapidly exploring Random Tree (RRT) algorithm which is efficient but also stochastic. For planning a path to the chosen frontier point, several path planning algorithms have been proposed. Bircher et al. \cite{article14} and Ellips and Hossein \cite{article20} make use of RRT, while Karaman and Frazoli \cite{article15} propose RRT* for path planning, a more efficient implementation of RRT. Stachniss et al. \cite{article21} uses a Rao-Blackwellized particle filter and Elhoseny et al. \cite{article22} proposes genetic algorithm-based path planning for the task. In this paper, we chose edge detection as the method for sampling frontier points and use A-star algorithm for the task of path planning. We do this to reduce the stochasticity in our results and reach definitive conclusions regarding the efficacy of our proposed heuristics.

Existing research has a variety of methods for choosing the heuristic function for Frontier planners. Yamauchi \cite{article5} selects the nearest frontier point and Topiwala et al. \cite{article16} use Breath First Search to select the next frontier point to explore. Simmons et al. \cite{article6} and Moorehead et al. \cite{article7} improve on this by using a combination of information gain and movement cost. 

Furthermore, reinforcing existing planning methods with future rewards can lead to improved results for exploration and coverage tasks, as demonstrated by \cite{article24} and \cite{article23}. To the authors’ best knowledge:
\begin{itemize}
\item The heuristics being used by  existing frontier planner methods do not account for the future reward that could be obtained from unexplored regions of the map.
\item The heuristic does not penalise the planner for overlooking sparse unexplored localities that incur high movement cost in the future.
\end{itemize}

In this paper, we propose a new heuristic for frontier planners that harnesses the ability of Deep Neural Networks to learn complex non-linear functions to predict future rewards. Neural Networks have been widely used for learning complex non-linear functions for a variety of tasks including heuristic learning. Takahashi et al. \cite{article18}, Qureshi et al. \cite{article13} and Ariki and Narihira \cite{article17} have demonstrated the efficiency of DNN's in learning heuristics for path planning tasks by using appropriate loss functions. Meanwhile Bhardwaj et al. \cite{article12} and Bency et al. \cite{article19} have provided an efficient algorithm for training networks to imitate clairvoyant oracles focusing on computing a heuristic that explicitly minimizes search effort. In addition to this, we also propose another addition to the heuristic function in the form of a filter based heuristic, that ensures a more efficient coverage of the given map.

\section{Methodology}
The state of the agent is represented using two 2D arrays: The map and the mask. The map is a 2D occupancy grid that stores the location of objects that have been detected by the agent during exploration. It stores a 1 for occupied cells and a 0 for unoccupied cells. The mask is used to represent the exploration information, storing a 1 for explored cells and a 0 for unexplored cells. The entire procedure used for the coverage of unknown environments is described in Algorithm \ref{algo}. \\

\subsection{Frontier Point Generations}

For this paper we have chosen to use edge detection techniques to generate frontier points. We use the Sobel filter and apply it on our mask to obtain the edges which represent the boundary between the explored and unexplored regions. As the map size increases, so do the number of frontier points, making the planner slower. To solve this we limit the number of frontier points considered for evaluation to $N$ points. The $N$ points are chosen by selecting the $k$ closest points and sampling the rest $N-k$ points from the list of remaining frontier points randomly.

\subsection{Path Planner}

We use A-Star as the path planner in this paper. We provide it with the map, the mask, the current location of the agent and a frontier point which acts as the target location. The planner considers the unexplored region to be unoccupied during the generation of the path. It returns the path from the current location to the target frontier point and a temporary mask of what our current mask would look like if we traversed the path.

\begin{algorithm}
\SetAlgoLined
\DontPrintSemicolon
 initialize $map$, $mask$ \;
 initialize $N$, $k$, $d$, $L$, $loc$\;
 \While{Unexplored\_Region != 0}{
  $edges \leftarrow Sobel(mask)$\;
  $fps \leftarrow sample\_points(edges,loc,N,k)$\;
  $H\_best \leftarrow -L^2$\;
  \For{each fp in fps}{
  $m, path \leftarrow a\_star(map,mask,loc, fp)$\;
  $H1 \leftarrow \sum_{i=0}^{L^2-1}\|mask_i-m_i\|$\;
  $H2 \leftarrow length(path)$\;
  $state \leftarrow local\_area(map, m, d, fp)$\;
  $H3 \leftarrow Network(state)$\;
  $H4 \leftarrow applyFilter(map, m, fp)$\;
  $H \leftarrow \alpha \times H1 - \beta \times H2 + \gamma \times H3 + \delta \times H4$\;
  \If{H\_best < H}{
  $H\_best \leftarrow H$\;
  $Best\_path \leftarrow path$\;
  }
%   {
%   $continue$\;
%   }
  }
  \For{each step in Best\_path}{
  $loc, mask, map \leftarrow agent.move(step)$
  }
  
 }
  \tcp{$m \leftarrow temp\_mask \in R^{d \times d}$}
  \tcp{$fps \leftarrow frontier\_points$}
  \tcp{$loc \leftarrow agent\_location$}
  \tcp{$L \leftarrow map\_length$}
 \caption{Frontier Based Coverage Algorithm}
 \label{algo}
\end{algorithm}

\subsection{Heuristic Function}

The heuristic used in this paper is a combination of the following heuristics:

\begin{itemize}
    
\item \textbf{Immediate Information Gain H1:} H1 represents the information gained while travelling from the agent's current location to a selected frontier point by traversing the path provided by the A-Star module. It is calculated by summing up the number of points that were initially unexplored but fell inside the scan radius of the agent during traversal of the path thus making them explored.
$$H1 = \sum_{i=0}^{L^2-1}\|mask_i - temp\_mask_i\|$$
where $L^2$ is the number of grid elements in the map, $mask$ is the current mask of the agent and $temp\_mask$ is the temporary mask the agent would have if it traversed the given path.

\item \textbf{Movement Cost H2:} H2 is the cost of moving from the current location to the selected frontier point. It is the number of steps in the path provided by the A-Star module.
$$H2 = P$$
where $P$ is the length of the path.

\item \textbf{Expected Future Return H3:} H3 is the gain in information for future steps taken from a selected frontier point. It is the summation of the H1 heuristic over future time steps.
$$H3 = \sum_{t=1}^{T}H1_t$$
where T is the total look-ahead time steps. It is computed by the network mentioned in sub-section D. 

\item \textbf{Custom Heuristic H4:} H4 is used to provide incentive to the planner to avoid leaving clusters of unexplored regions surrounded by explored regions. It is computed using a set of filters that are applied on $temp\_mask$ with the filter centred at the selected frontier point. Each filter is coupled with an activation threshold $\xi$, when the value returned by bit-wise multiplying the filter with the region centred at the selected frontier point is higher than $\xi$ the heuristic gets activated. On activation the value of H4 is set to 1. The filters are shown in Fig. \ref{fig:test1}. The main heuristic function is as follows:
$$H = \alpha \times H1 - \beta \times H2 + \gamma \times H3 + \delta \times H4$$

The filters, when applied on the $temp\_mask$, return the number of explored gird cells in that region. The thresholds decide the acceptable number of explored cells for the heuristic to activate and as the filter gets larger the acceptable number of unexplored cells increases.
% $
% \begin{bmatrix}
% 1 & 1 & 1 & 1 & 1\\
% 1 & 1 & 1 & 1 & 1\\
% 1 & 1 & 0 & 1 & 1\\
% 1 & 1 & 1 & 1 & 1\\
% 1 & 1 & 1 & 1 & 1\\
% \end{bmatrix}
% $

\end{itemize}

% \begin{figure*}
% \begin{tabular}{cc}
% \subfloat[Filters]{\framebox{\parbox{3in}{
% $F_1 = \begin{bmatrix}
% 1 & 1 & 1 & 1 & 1\\
% 1 & 1 & 1 & 1 & 1\\
% 1 & 1 & 0 & 1 & 1\\
% 1 & 1 & 1 & 1 & 1\\
% 1 & 1 & 1 & 1 & 1\\
% \end{bmatrix} \quad \xi_1 = 24
% $

% $\quad \\$

% $F_2 = \begin{bmatrix}
% 1 & 1 & 1 & 1 & 1 & 1 & 1\\
% 1 & 1 & 1 & 1 & 1 & 1 & 1\\
% 1 & 1 & 1 & 1 & 1 & 1 & 1\\
% 1 & 1 & 1 & 0 & 1 & 1 & 1\\
% 1 & 1 & 1 & 1 & 1 & 1 & 1\\
% 1 & 1 & 1 & 1 & 1 & 1 & 1\\
% 1 & 1 & 1 & 1 & 1 & 1 & 1\\
% \end{bmatrix} \quad \xi_2 = 47
% $

% $\quad \\$

% $F_3 = \begin{bmatrix}
% 1 & 1 & 1 & 1 & 1 & 1 & 1 & 1 & 1\\
% 1 & 1 & 1 & 1 & 1 & 1 & 1 & 1 & 1\\
% 1 & 1 & 1 & 1 & 1 & 1 & 1 & 1 & 1\\
% 1 & 1 & 1 & 1 & 1 & 1 & 1 & 1 & 1\\
% 1 & 1 & 1 & 1 & 0 & 1 & 1 & 1 & 1\\
% 1 & 1 & 1 & 1 & 1 & 1 & 1 & 1 & 1\\
% 1 & 1 & 1 & 1 & 1 & 1 & 1 & 1 & 1\\
% 1 & 1 & 1 & 1 & 1 & 1 & 1 & 1 & 1\\
% 1 & 1 & 1 & 1 & 1 & 1 & 1 & 1 & 1\\
% \end{bmatrix} \space \xi_3 = 78
% $

% }}} &
% \subfloat[Network]{\includegraphics[scale=0.27]{images/model_2.png}}
% \end{tabular}
% \end{figure*}

\subsection{Training Procedure}

\begin{figure}
\centering
\begin{minipage}{.5\textwidth}
  \centering
  \framebox{\parbox{3in}{
$F_1 = \begin{bmatrix}
1 & 1 & 1 & 1 & 1\\
1 & 1 & 1 & 1 & 1\\
1 & 1 & 0 & 1 & 1\\
1 & 1 & 1 & 1 & 1\\
1 & 1 & 1 & 1 & 1\\
\end{bmatrix} \quad \xi_1 = 24
$

$\quad \\$

$F_2 = \begin{bmatrix}
1 & 1 & 1 & 1 & 1 & 1 & 1\\
1 & 1 & 1 & 1 & 1 & 1 & 1\\
1 & 1 & 1 & 1 & 1 & 1 & 1\\
1 & 1 & 1 & 0 & 1 & 1 & 1\\
1 & 1 & 1 & 1 & 1 & 1 & 1\\
1 & 1 & 1 & 1 & 1 & 1 & 1\\
1 & 1 & 1 & 1 & 1 & 1 & 1\\
\end{bmatrix} \quad \xi_2 = 47
$

$\quad \\$

$F_3 = \begin{bmatrix}
1 & 1 & 1 & 1 & 1 & 1 & 1 & 1 & 1\\
1 & 1 & 1 & 1 & 1 & 1 & 1 & 1 & 1\\
1 & 1 & 1 & 1 & 1 & 1 & 1 & 1 & 1\\
1 & 1 & 1 & 1 & 1 & 1 & 1 & 1 & 1\\
1 & 1 & 1 & 1 & 0 & 1 & 1 & 1 & 1\\
1 & 1 & 1 & 1 & 1 & 1 & 1 & 1 & 1\\
1 & 1 & 1 & 1 & 1 & 1 & 1 & 1 & 1\\
1 & 1 & 1 & 1 & 1 & 1 & 1 & 1 & 1\\
1 & 1 & 1 & 1 & 1 & 1 & 1 & 1 & 1\\
\end{bmatrix} \space \xi_3 = 78
$

}}
  \caption{Filters $F_1$, $F_2$ and $F_3$ along with their corresponding activation thresholds $\xi_1$, $\xi_2$ and $\xi_3$}
  \label{fig:test1}
\end{minipage}%
\begin{minipage}{.5\textwidth}
  \centering
  \includegraphics[width=.4\linewidth]{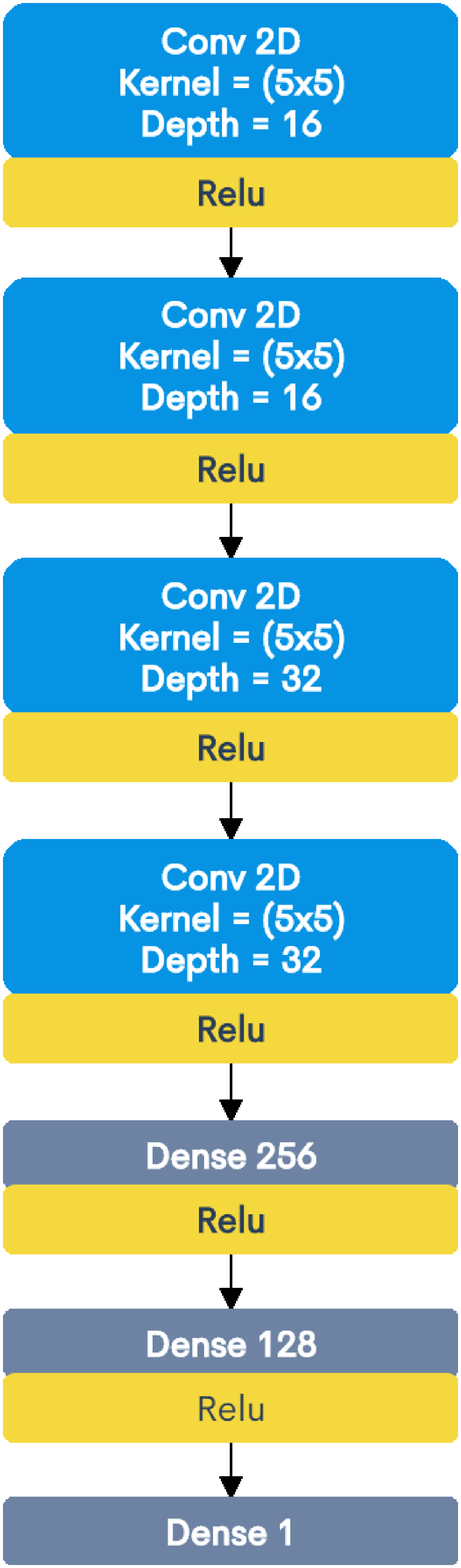}
  \caption{Architecture of heuristic network}
  \label{fig:test2}
\end{minipage}
\end{figure}

\begin{itemize}

\item \textbf{Network:} The network is a CNN that uses local information with respect to the frontier point. We extract $d \times d$ regions centered about the location of the frontier point from both the $temp\_mask$ and the $map$ which are resized to $50 \times 50$ grids before being stacked together to obtain the network input state.
% $$F_\theta : x \rightarrow R \quad ; \quad x \in R^{[50 \times 50 \times 3]}$$

$$
min_{\theta' \in \theta} \space \xi(f_{\theta'}) \space ; \quad {\xi}({f_{\theta'}}) = \frac{1}{n}\sum_i^n \mathbb{L}(f_{\theta'}(x),y)
$$

Where $\theta'$ represents the parameters of the network. The output of the network is a real number which represents the expected future return. The architecture of the network is shown in Fig. \ref{fig:test2}. and the loss $\mathbb{L}$ used to train the network is Mean Squared Error (MSE).

% \item \textbf{Network:} The network is a CNN that takes a local area of the $temp\_mask$ and the $map$ centred at the location of the selected frontier point. The input is a $d \times d \times 2$ image where the first channel is a $d \times d$ section of the $map$ and the second channel is a $d \times d$ section of the $temp\_mask$.
% $$F_\theta : x \rightarrow R \quad ; \quad x \in R^{[d \times d \times 2]}$$
% Where $\theta$ represents the parameters of the network. The output of the network is a real number which represents the expected future return. The architecture of the network is shown in Fig. \ref{network}. and the loss used to train the network is Mean Squared Error (MSE).
\item \textbf{Oracle:} The oracle is responsible for providing the labels for the training data provided to the network. It is computed by building a tree of future states and using it to generate the expected future reward. The oracle has access to the unexplored regions of the environment allowing it to provide accurate estimates of future reward.
\newline

\end{itemize}

\section{Results and Experiments}

\subsection{Experimental Set-up}

The environment used for the experiments is a grid world with a fixed sensor radius for scanning the surroundings and no sensory noise has been added for the experiments. The maps used for testing the planner were generated using a random map generator script that generates $L \times L$ sized maps with random placement of obstacles of various sizes. By allowing the generation process to be random we are eliminating map based biases allowing the evaluation metrics used to be bias free. The network was implemented using Tensorflow and the environment along with the rest of the modules can be found here. In the experiments conducted in this section, we've set $N=15$ and $k=12$ for the sampling of the frontier points. We have also chosen the thresholds $\xi_1 = 24$, $\xi_2 = 47$ and $\xi_3 = 78$ for the filters in the H4 heuristic after careful consideration of the conditions in which this heuristic should be activated and by experimenting and fine-tuning these thresholds to match the desired output.

\begin{figure*}
\begin{tabular}{cccc}
\subfloat[Step = 400]{\includegraphics[width = 1.3in]{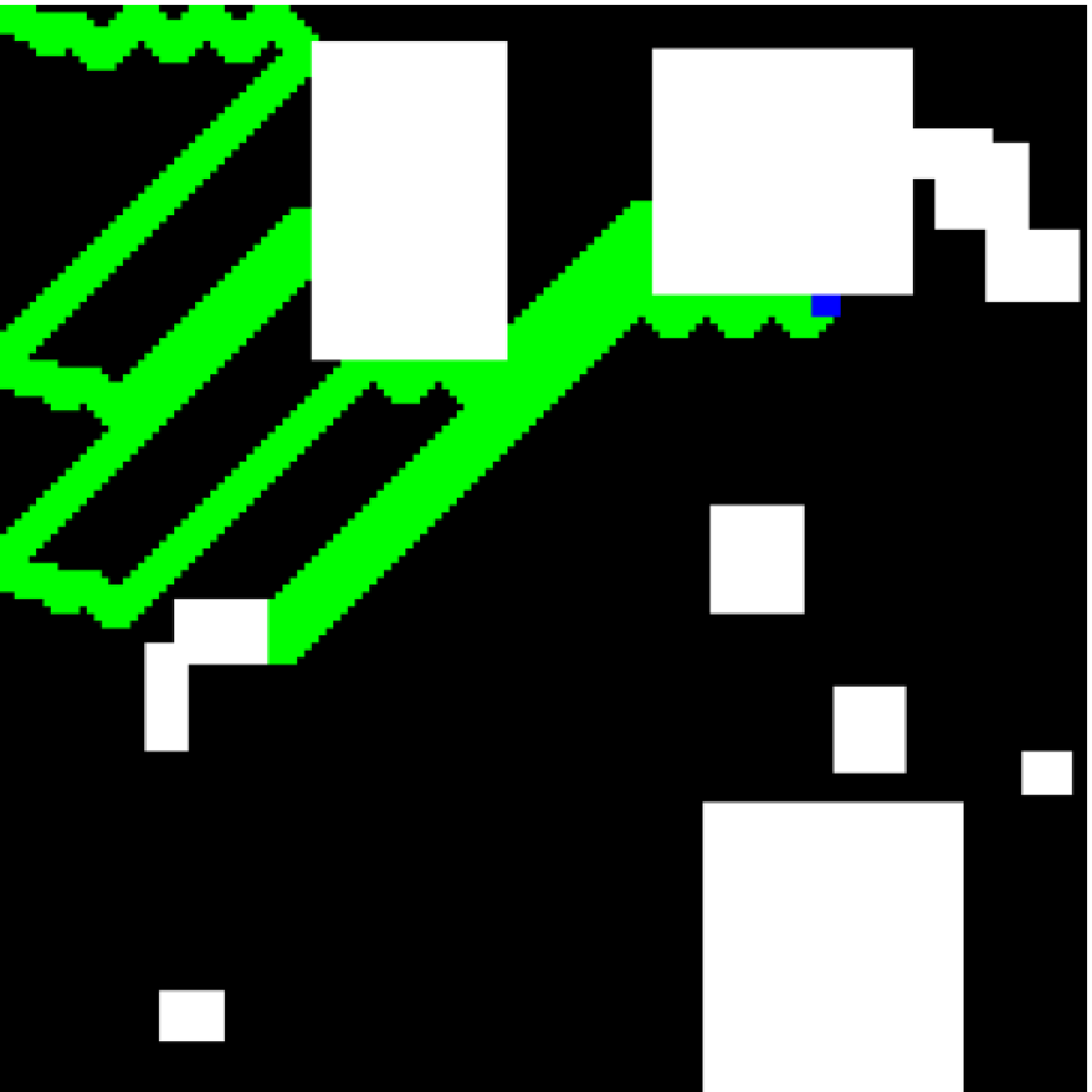}} &
\subfloat[Step = 1500]{\includegraphics[width = 1.3in]{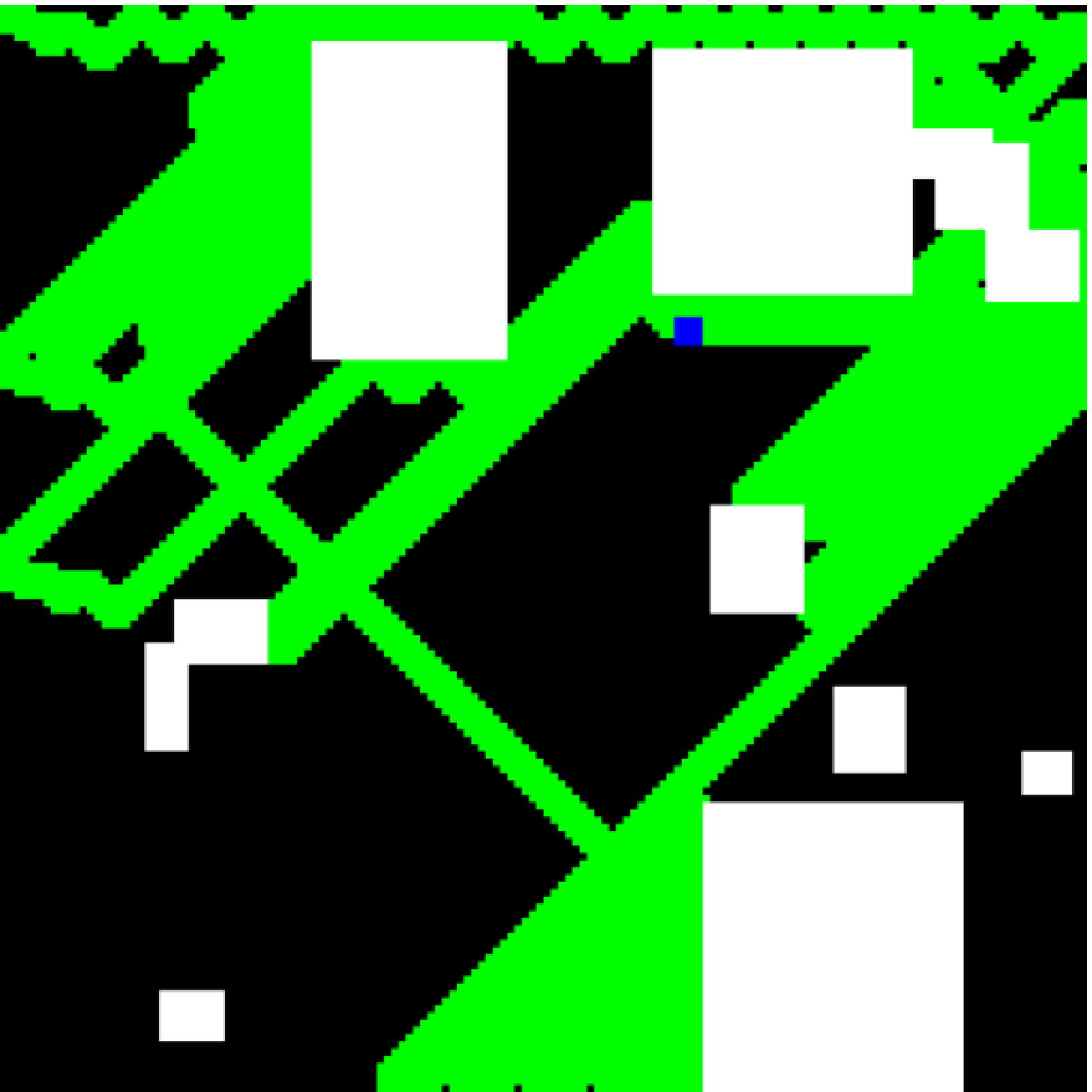}} &
\subfloat[Step = 3000]{\includegraphics[width = 1.3in]{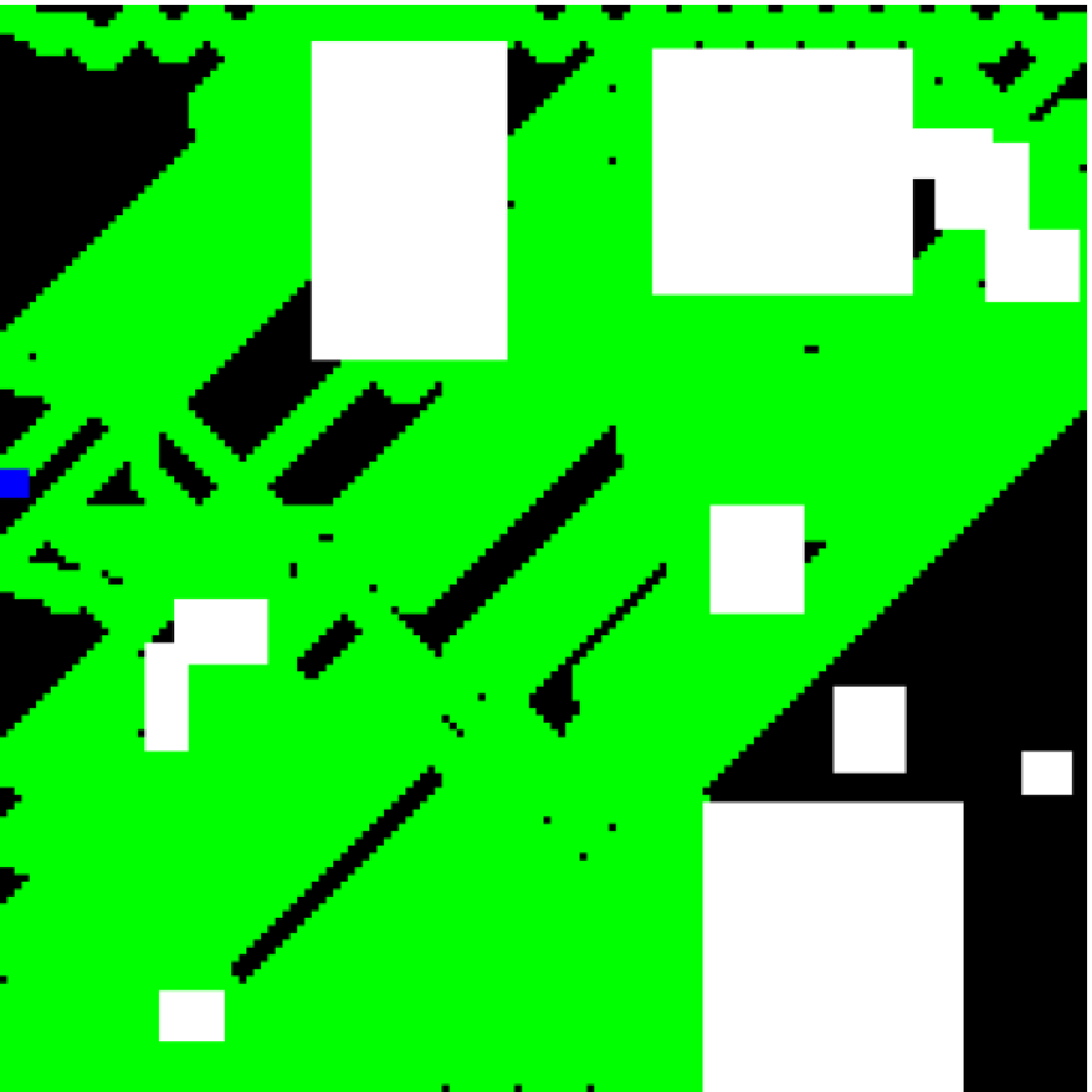}} &
\subfloat[Step = 4500]{\includegraphics[width = 1.3in]{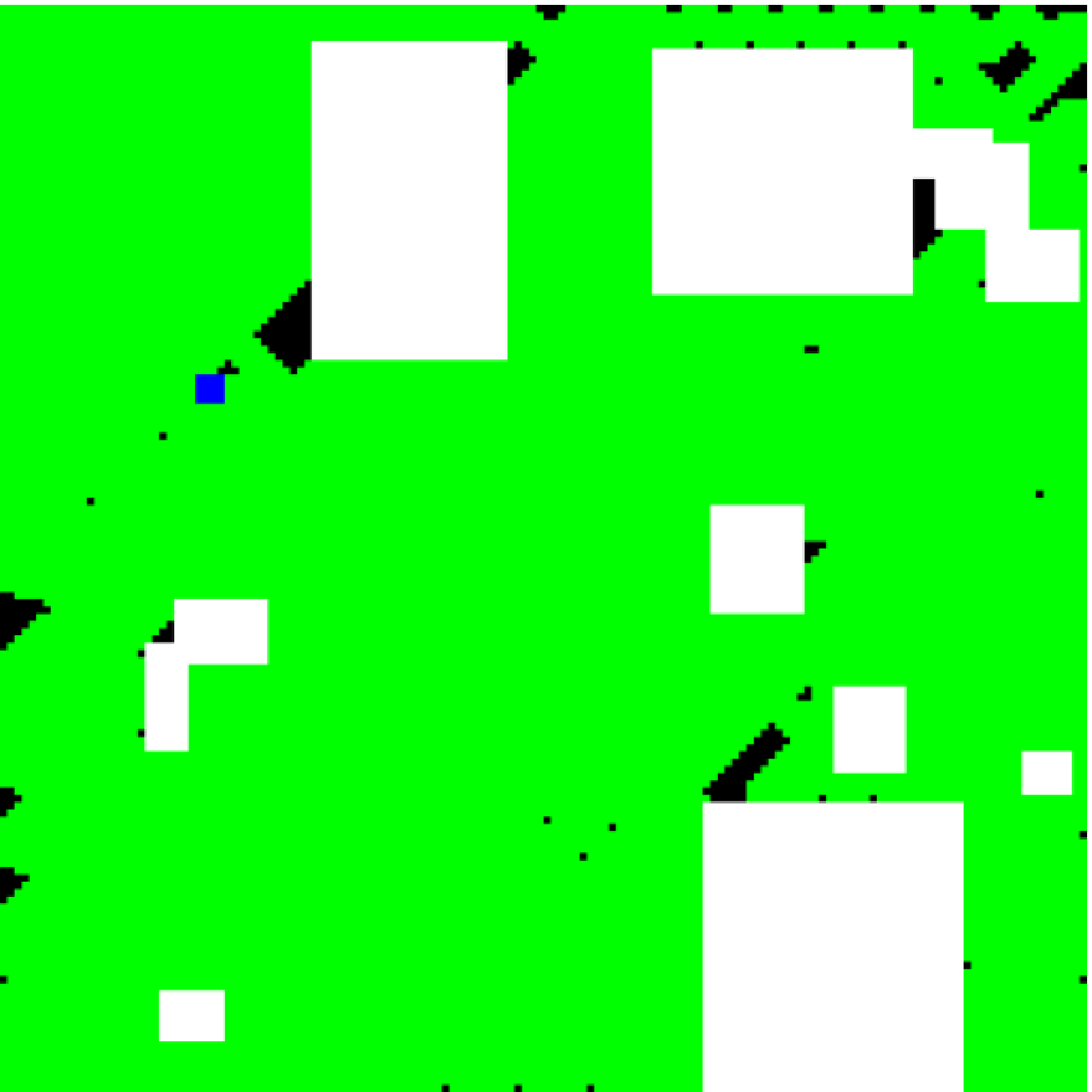}}\\
\subfloat[Step = 400]{\includegraphics[width = 1.3in]{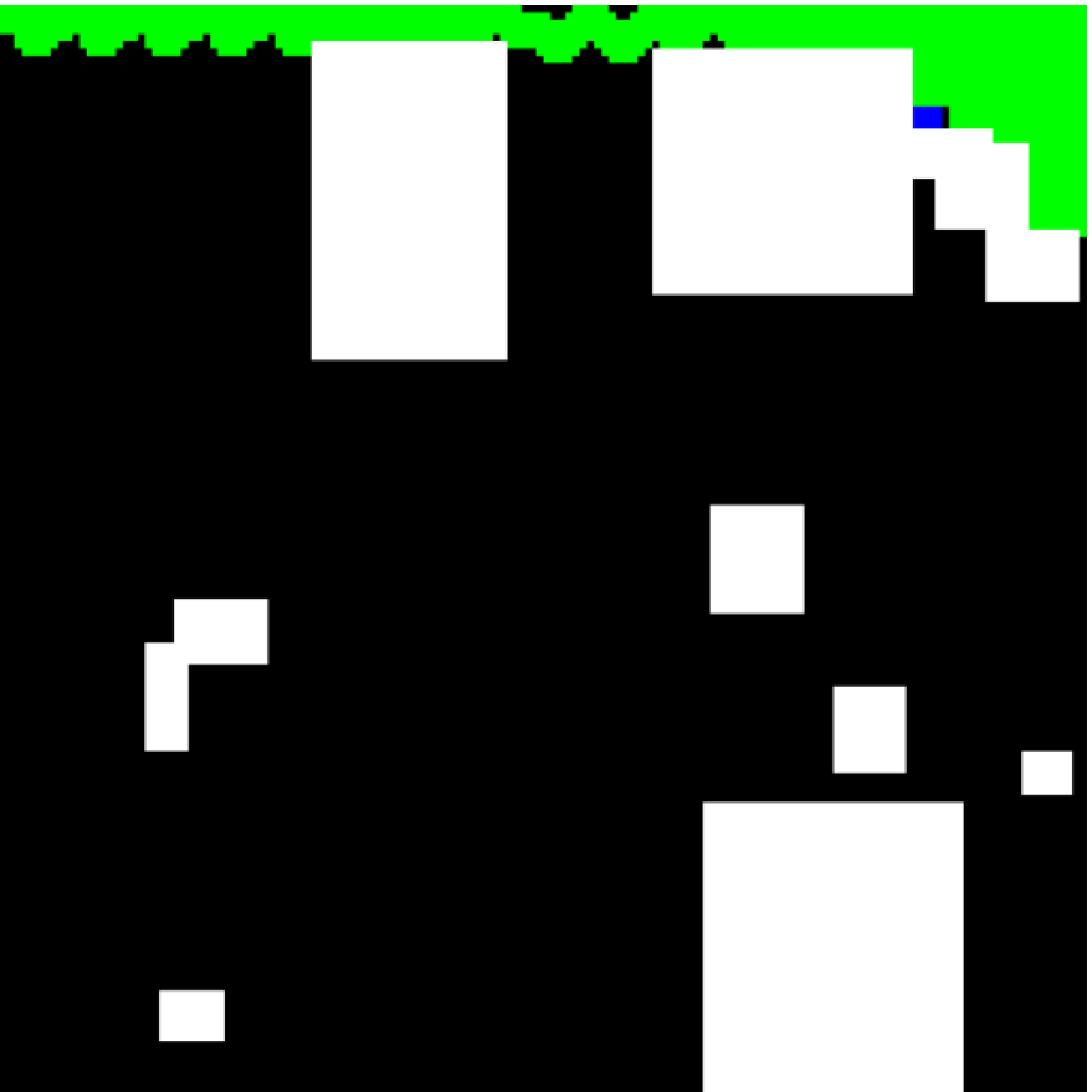}} &
\subfloat[Step = 1500]{\includegraphics[width = 1.3in]{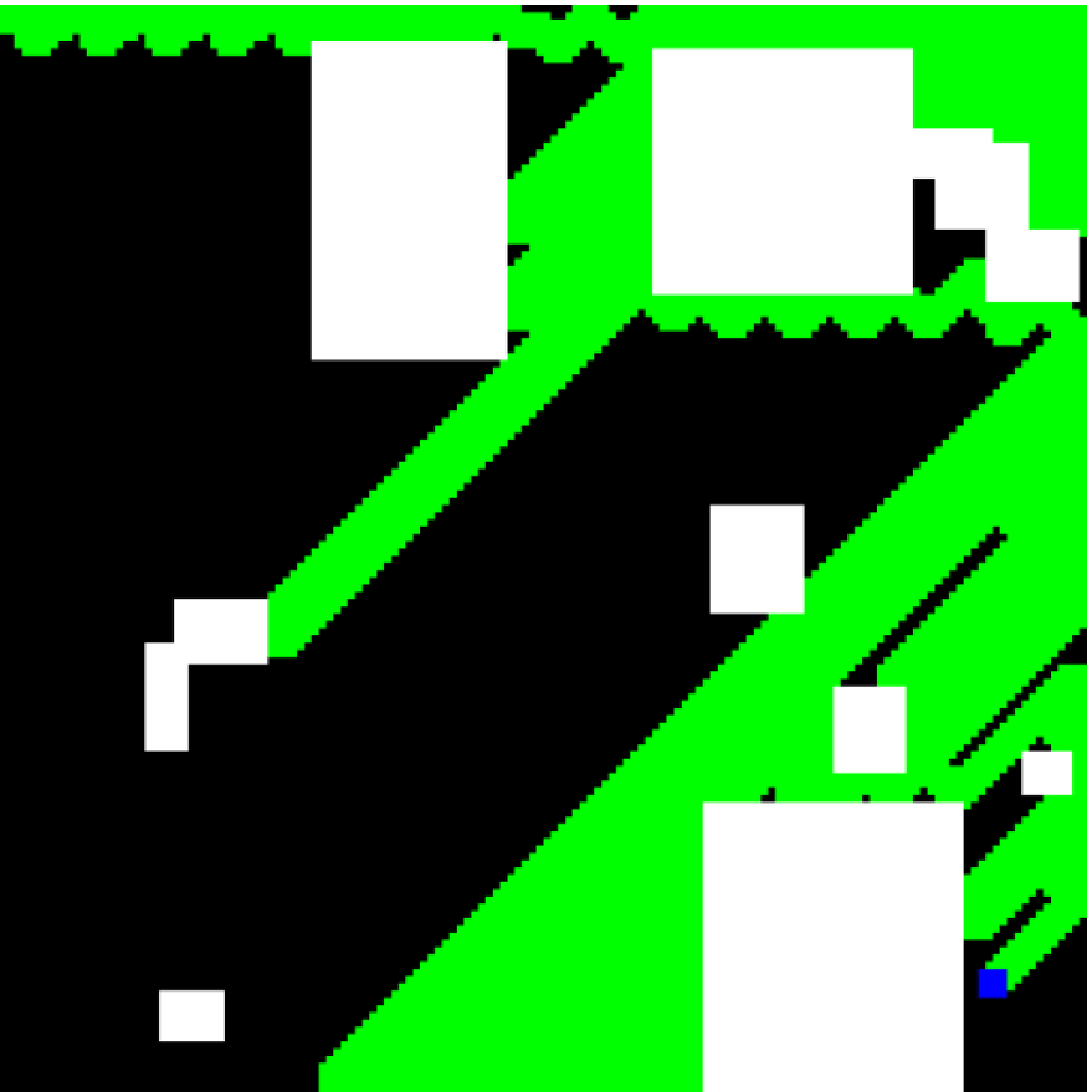}} &
\subfloat[Step = 3000]{\includegraphics[width = 1.3in]{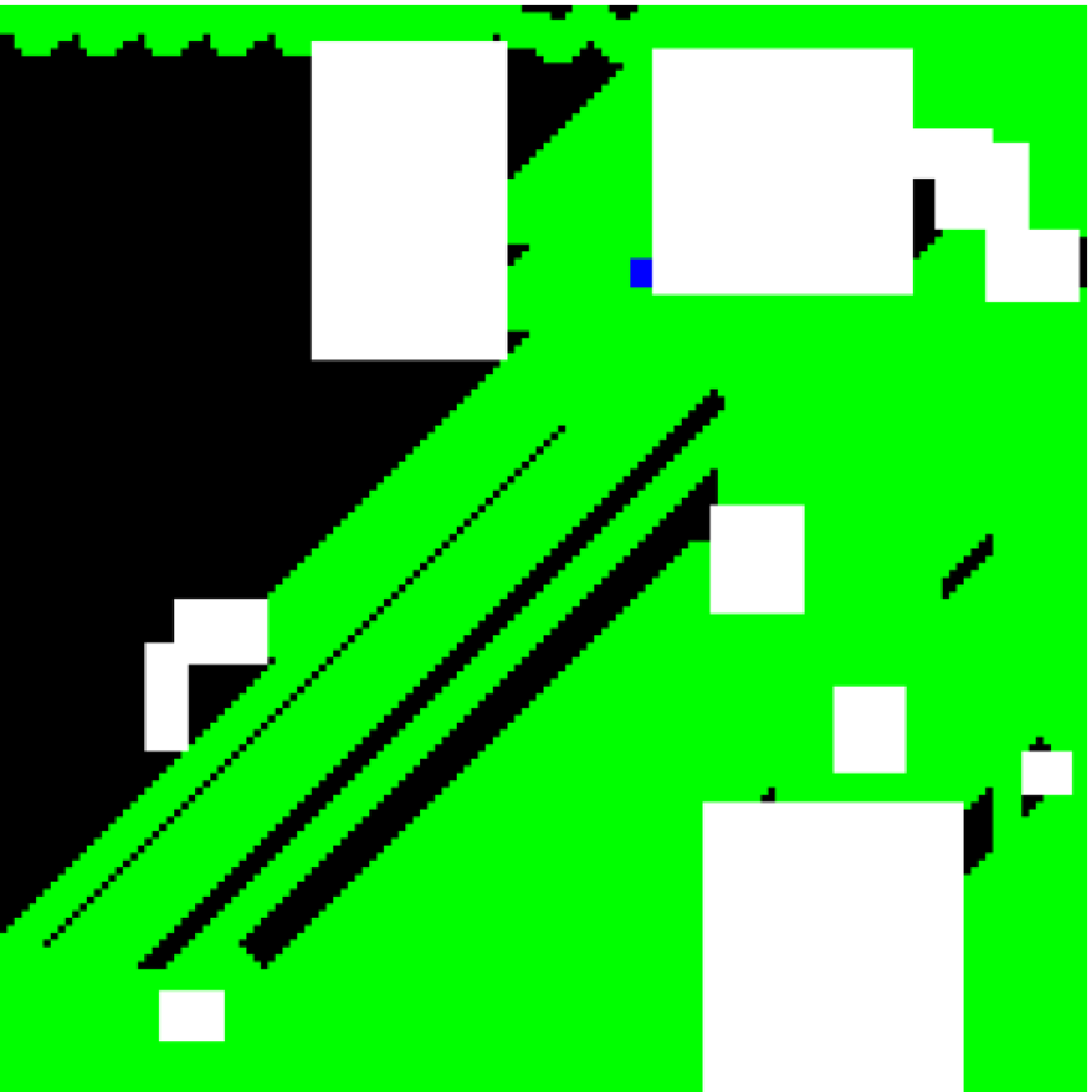}} &
\subfloat[Step = 4500]{\includegraphics[width = 1.3in]{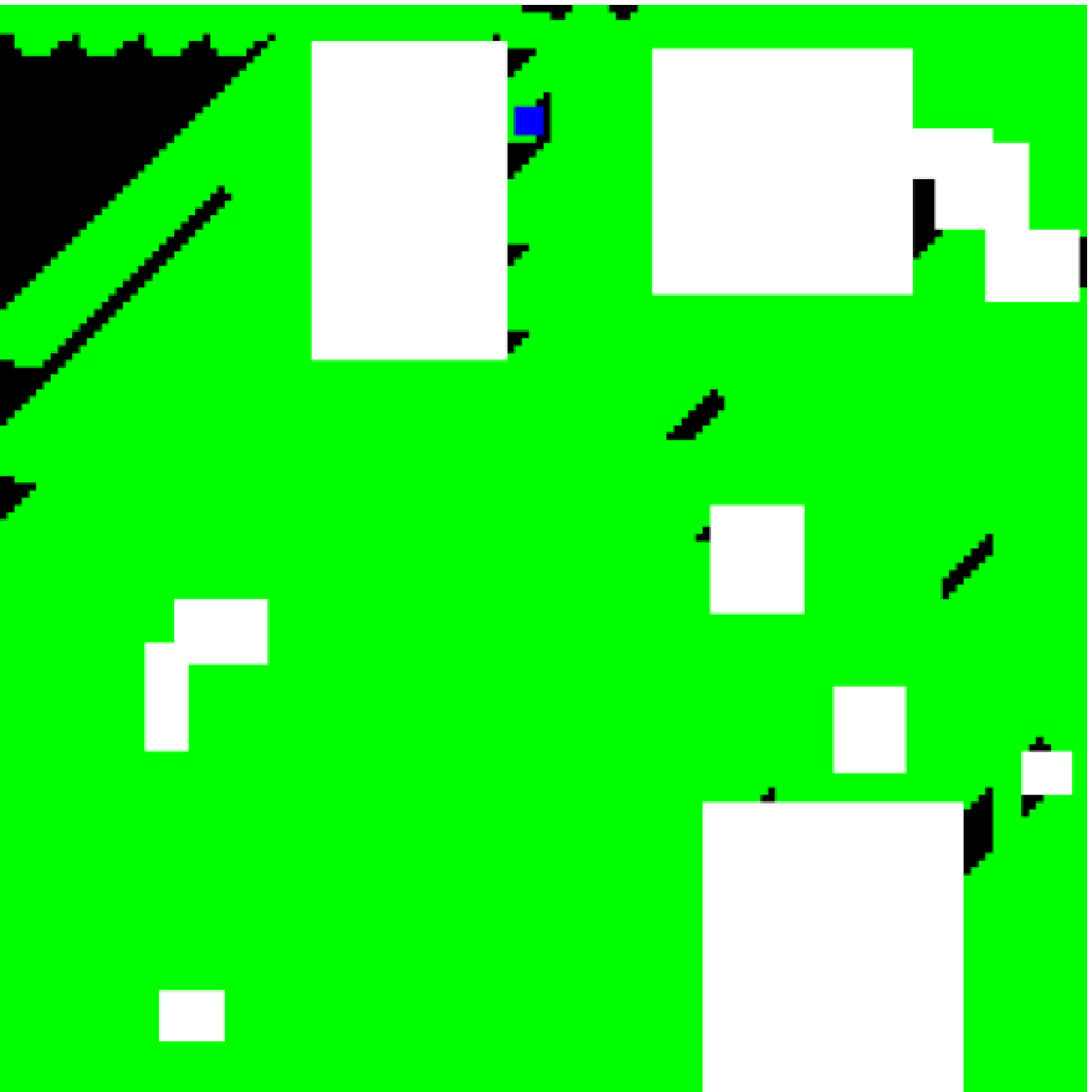}}
\end{tabular}
\caption{Visualisation of the head to head comparison between the planner with and without the use of the proposed heuristics. Images (a), (b), (c) and (d) belong to the episode without the use of H3 and H4 which completed the map in 6163 steps while images (e), (f), (g) and (h) belong to the the episode with the use of H3 and H4 which completed the map in 5946 steps. The white regions represent occupied cells, the black regions represent unexplored unoccupied cells, the green region represents the explored unoccupied cells and the blue square is the location of the agent.}
\end{figure*}

\subsection{Metrics}

The metric chosen for evaluating the performance of the heuristic is the number of steps required to build a complete map of the environment. The stochastic nature of the frontier point sampling process requires us to run multiple episodes on the same environment and take the final number of steps taken to build a map of the environment as the mean of the number of steps taken per episode. We evaluate the heuristic on several maps and the final metric $\mu$ is the mean of the number of steps for a given map over all the maps. \\

\begin{center}
$
\mu = \frac{1}{N_M} \sum_{i=1}^{N_M} (\frac{1}{N_E} \sum_{j=1}^{N_E} steps_{j}^{i})
$ \\
\end{center}

Where $N_M$ is the number of maps, $N_E$ is the number of episodes per map and $steps_{j}^{i}$ is the number of steps taken to build a map of the environment for $map_i$, $episode_j$.

The variance $\sigma$ is also reported for the conducted experiments and it denotes the variance in the number of steps for the given maps. Lower variance signifies consistency in performance of the planner for the given set of parameters. \\

\begin{center}
$
\sigma = \frac{1}{N_M} \sum_{i=1}^{N_M} (\frac{1}{N_E} \sum_{j=1}^{N_E} steps_{j}^{i})^2 - \mu^2
$ \\
\end{center}

The parameters that can be tuned are the heuristic coefficients $\alpha$, $\beta$, $\gamma$, $\delta$, the network state size $d$, and the look-ahead $T$. In this work, the parameter tuning process starts with the heuristic coefficients $\alpha$, $\beta$ and $\gamma$. We then tune the state size parameter $d$, followed by the look-ahead parameter $T$. Lastly we tune the H4 heuristic parameter $\delta$ and evaluate the performance gain for the planner with respect to the final heuristic function. The parameters were tuned on a set of 28 randomly generated $50 \times 50$ maps and the planner was run for 10 episodes on each map for a given set of parameters.

\subsection{Heuristic Coefficients $\alpha$, $\beta$ and $\gamma$}

The coefficients $\alpha$, $\beta$ and $\gamma$ were tuned by fixing the state size parameter $d$ to 20 and setting the look-ahead parameter $T$ as 1. Table \ref{param_table} shows the head to head performance of the planner with and without the use of the expected future return H3 for various values of $\alpha$ and $\beta$. In our experiments we found that the best performance was observed when $\alpha$ and $\gamma$ were similar, hence we have set both $\alpha$ and $\gamma$ as 1 for the experiments in Table \ref{param_table}. The best parameters for the planner without the use of H3 is found to be $\alpha:\beta = 1:15$ while the  best parameters with the use of H3 are found to be $\alpha:\beta:\gamma = 1:12:1$.

\subsection{Network State Size $d$}

The network state size $d$ decides the size of the local window of the map and the mask provided to the network to learn the H3 heuristic. Table \ref{table_state} shows the performance of the planner for different values of $d$. The heuristic coefficients are set as $\alpha:\beta:\gamma = 1:12:1$ and the look-ahead $T$ is set as 1. The performance is observed to be the best for $d$ equal to 40.

\subsection{Look-Ahead $T$}

The Look-Ahead $T$ is the total number of steps into the future we make our network predict. The oracle builds a tree of depth $T$ for providing the labels for the training process. The performance of the network with respect to $T$ is analyzed in Table \ref{table_look-ahead}. We fix $\alpha:\beta:\gamma = 1:12:1$ and these parameters are coupled with various combinations of $d$ and $T$. In our experiments we observe that larger values of $d$ provide better performance irrespective of the value of $T$. Hence we show the comparison of different values of $T$ for $d=40$ in Table \ref{table_look-ahead} and the best performance for the planner is observed when $T$ is set to 1.

\subsection{Heuristic Coefficient $\delta$}

The coefficient $\delta$ controls the contribution of H4 to the heuristic function $H$. It was tuned by setting $\alpha:\beta:\gamma = 1:12:1$, $d=40$ and $T = 1$. The comparison of different values of H4 is given in Table \ref{H4_table}. $\delta$ was noted to provide the best performance when set to 20.

\begin{table}
\parbox{.45\linewidth}{
\centering
\begin{tabular}{|c|c|c|c|c|c|}
\hline
\multirow{2}{*}{$\alpha$} & \multirow{2}{*}{$\beta$} & \multicolumn{2}{c|}{Mean $\mu$} & \multicolumn{2}{c|}{Variance $\sigma$}\\
% \hline

\cline{3-6}

& & $\gamma$ = 0 & $\gamma$ = 1 & $\gamma$ = 0 & $\gamma$ = 1\\
\hline
3 & 1 & 1076 & 1086 & 17565 & 15850\\
\hline
1 & 1 & 917 & 934 & 9832 & 9127\\
\hline
1 & 5 & 723 & 730 & 4945 & 5070\\
\hline
1 & 10 & 699 & 684 & 4266 & 4728\\
\hline
\textbf{1} & \textbf{12} & \textbf{691} & \textbf{675} & \textbf{5080} & \textbf{4798}\\
\hline
1 & 15 & 690 & 682 & 4917 & 4255\\
\hline
1 & 20 & 691 & 680 & 4500 & 4630\\
\hline
\end{tabular}
\vspace*{2mm}
\caption{Performance Under Parameter Variation for $50 \times\ 50$ Maps}
\label{param_table}
}
\hfill
\parbox{.45\linewidth}{
\centering
\begin{tabular}{|c|c|c|}
\hline
State ($d \times d$) & Mean $\mu$ &  Variance $\sigma$\\
\hline
20x20 & 675 & 4798\\
\hline
30x30 & 669 & 5453\\
\hline
\textbf{40x40} & \textbf{667} & \textbf{4291}\\
\hline
\end{tabular}
\vspace*{2mm}
\caption{Comparison of Performance for different State size $d \times d$}
\label{table_state}
}

\hfill
\parbox{.45\linewidth}{
\vspace*{0.5mm}
\hspace*{-7mm}
\centering
\begin{tabular}{|c|c|c|c|}
\hline
look-ahead $T$ & MSE & Mean $\mu$ &  Variance $\sigma$\\
\hline
\textbf{1} & \textbf{2.62} & \textbf{667} & \textbf{4291}\\
\hline
2 & 8.1 & 677 & 4456\\
\hline
3 & 40.26 & 693 & 4706\\
\hline
\end{tabular}
\vspace*{2mm}
\hspace*{-15mm}
\caption{Comparison of look-ahead $T$ for H3}
\label{table_look-ahead}
}
\hfill
\parbox{.45\linewidth}{
\centering
\vspace{5mm}\begin{tabular}{|c|c|c|}
\hline
$\delta$ & Mean $\mu$ &  Variance $\sigma$\\
\hline
10 & 659 & 3870\\
\hline
15 & 650 & 3991\\
\hline
\textbf{20} & \textbf{645} & \textbf{3296}\\
\hline
25 & 651 & 4304\\
\hline
30 & 649 & 3990\\
\hline
\end{tabular}
\vspace*{2mm}
\caption{Performance Comparison For different values of $\delta$}
\label{H4_table}
}
\end{table}

\begin{figure}

\begin{tikzpicture}
% \caption{Demo}
\begin{axis}[
    title={Performance Analysis},
    xlabel={Map Size L},
    ylabel={Performance Gain S (\%)},
    xmin=40, xmax=160,
    ymin=0, ymax=10,
    xtick={50,75,100,125,150},
    ytick={1.0,2.0,3.0,4.0,5.0,6.0,7.0,8.0,9.0,10.0},
    legend pos=north west,
    ymajorgrids=true,
    grid style=dashed,
]

\addplot[
    color=blue,
    mark=square,
    ]
    coordinates {
    (50,3.6)(75,4)(100,3.48)(150,2.52)
    };
    % \legend{c = 0}

\addplot[
    color=red,
    mark=square,
    ]
    coordinates {
    (50,6.52)(75, 6.21)(100, 7.12)(150, 6.7)
    };
    \legend{$\gamma$ = 1  $\delta$ = 0,$\gamma$ = 1  $\delta$ = 20}
    
\end{axis}
\end{tikzpicture} \\

\scalebox{0.9}{
\begin{tabular}{|c|c|c|c|c|}
\hline
Map Size & 50x50 & 75x75 &  100x100 & 150x150\\
\hline
$S_{\gamma' = 1, \delta' = 0}$ & 3.6\% & 4.0\% & 3.4\% & 2.5\%\\
\hline
$S_{\gamma' = 1, \delta' = 20}$ & 6.5\% & 6.2\% & 7.1\% & 6.7\%\\
\hline
\end{tabular}} \\

\caption{Analysis of metric $S$ for the planner equipped with different heuristic parameters $\gamma$ and $\delta$ for increasing map sizes.}
\label{plot1}

\end{figure}

\subsection{Performance Analysis}

In this sub-section we compare the performance of the planner with the use of the proposed heuristics H3 and H4 alongside the performance of the planner without the use of the proposed heuristics. We use the following metric to compare the performance of a given heuristic function:
$$
S = \frac{\mu_{\gamma=0, \delta=0} - \mu_{\gamma=\gamma', \delta=\delta'}}{\mu_{\gamma=0, \delta=0}} \times 100
$$

$S$ represents the improvement in $\mu$ as a percentage. The parameters used for the expirements in Fig. \ref{plot1} are the parameters found to provide the best performance in the experiments conducted in the previous sections. $\alpha:\beta = 1:15$ for the heuristic function of the planner without the use of H3 and H4. $\alpha:\beta:\gamma=1:12:1$ is used in the heuristic function for the planner with the use of H3 and $\alpha:\beta:\gamma:\delta = 1:12:1:20$ are the parameters used in the heuristic function for the planner with the use of both H3 and H4. The state size is set as $d=40$ and the look-ahead for the network is set as $T = 1$.

In Fig. \ref{plot1} we plot the value of $S$ of the planner with the use of H3 only alongside the performance of the planner with the use of both H3 and H4 for different map sizes. We use 20 randomly generated maps for each of the different map sizes in this experiment and run each map for 10 episodes. The planner with the use of H3 performs 3.37\% better on average and the planner with the use of both H3 and H4 performs 6.62\% better on average.

In our experiments, we observe that the performance of the planner when equipped with H3 and H4 slightly diminishes for certain types of highly populated dense maps, however the performance gets sharply boosted for moderately populated and sparse maps giving us a better performance on average while using H3 and H4 in our heuristic function. This can be clearly observed in Fig. \ref{plot1}, where the planner with the use of H3 alone and the planner with the use of both H3 and H4 consistently perform better than the planner without the proposed heuristics on various map sizes. This is further substantiated by the visualisation of the head to head comparison of the planner with and without the use of H3 and H4 in Fig. 3, which exhibits that using the proposed heuristics prevents large clusters of unexplored area being left behind and leads to the agent following a Boustrophedon-like\cite{article25} policy, resulting in a more structured coverage of the map.

\section{Conclusion And Future Work}

In this paper, we have proposed two new heuristics which result in consistent improvements in the performance of frontier planners for the task of efficient coverage of unknown environments. We leverage the ability of deep neural networks to learn complicated non-linear functions to learn auxiliary heuristics and use this alongside custom filter-based heuristics to demonstrate our claim.

Future work could include the use of probabilistic graphical networks that accept undirected graphs of the entire map as inputs as opposed to fixed local areas of the map to allow the network to make predictions based on global information as opposed to local information. The algorithm could also be adapted to work for multi-agent scenarios by allowing the network to allocate frontier points to each agent with respect to the other agents.  

\section{Additional information}
\subsection{Notes on contributors}
Arsh Tangri: Arsh Tangri is currently in the fourth year of his B.Tech in Electronics and Communication from the Manipal Institute of Technology, Manipal. His research interests are in the areas of Reinforcement Learning, Deep Learning and Robotics. \newline \newline Dhruv Joshi: Dhruv Joshi is currently in the fourth year of his B.Tech in Computer Science from the Manipal Institute of Technology, Manipal. His research interests are in the areas of Reinforcement Learning, Deep Learning and Robotics. \newline \newline Dr.Ashalatha Nayak is currently a Professor and the Head of Department for the Department of Computer Science in Manipal Insitute of Technology. Her research interesets are in Machine Learning, Computer Engineering and Model Based Testing

\section*{ACKNOWLEDGMENT}

We thank Project MANAS, Manipal for supporting us with the necessary resources.

\nocite{*}
\bibliographystyle{tfnlm}
\bibliography{bib_file}

\begin{thebibliography}{10}
\providecommand{\url}[1]{\normalfont{#1}}
\providecommand{\urlprefix}{Available from: }

\bibitem{article2}
Keidar~M, Kaminka~GA. Efficient frontier detection for robot exploration. The
  International Journal of Robotics Research. 2014;\hspace{0pt}33(2):215--236.

\bibitem{article3}
{Senarathne}~PGCN, {Wang}~D, {Wang}~Z, et~al. Efficient frontier detection and
  management for robot exploration. In: 2013 IEEE International Conference on
  Cyber Technology in Automation, Control and Intelligent Systems; 2013. p.
  114--119.

\bibitem{article4}
{Umari}~H, {Mukhopadhyay}~S. Autonomous robotic exploration based on multiple
  rapidly-exploring randomized trees. In: 2017 IEEE/RSJ International
  Conference on Intelligent Robots and Systems (IROS); 2017. p. 1396--1402.

\bibitem{article14}
Bircher~A, Kamel~MS, Alexis~K, et~al. Receding horizon "next-best-view" planner
  for 3d exploration. 05; 2016. p. 1462--1468.

\bibitem{article20}
Masehian~E, Kakahaji~H. Nrr: a nonholonomic random replanner for navigation of
  car-like robots in unknown environments. Robotica.
  2014;\hspace{0pt}32(7):1101–1123.

\bibitem{article15}
Karaman~S, Frazzoli~E. Sampling-based algorithms for optimal motion planning ;
  2011.

\bibitem{article21}
Stachniss~C, Grisetti~G, Burgard~W. Information gain-based exploration using
  rao-blackwellized particle filters. 06; 2005. p. 65--72.

\bibitem{article22}
Elhoseny~M, Shehab~A, Yuan~X. Optimizing robot path in dynamic environments
  using genetic algorithm and bezier curve. Journal of Intelligent \&\ Fuzzy
  Systems. 2017 09;\hspace{0pt}33:2305--2316.

\bibitem{article5}
{Yamauchi}~B. A frontier-based approach for autonomous exploration. In:
  Proceedings 1997 IEEE International Symposium on Computational Intelligence
  in Robotics and Automation CIRA'97. 'Towards New Computational Principles for
  Robotics and Automation'; 1997. p. 146--151.

\bibitem{article16}
Topiwala~A, Inani~P, Kathpal~A. Frontier based exploration for autonomous robot
  ; 2018.

\bibitem{article6}
Simmons~R, Apfelbaum~D, Burgard~W, et~al. Coordination for multi-robot
  exploration and mapping. 01; 2000. p. 852--858.

\bibitem{article7}
{Moorehead}~SJ, {Simmons}~R, {Whittaker}~WL. Autonomous exploration using
  multiple sources of information. In: Proceedings 2001 ICRA. IEEE
  International Conference on Robotics and Automation (Cat. No.01CH37164);
  Vol.~3; 2001. p. 3098--3103 vol.3.

\bibitem{article24}
{Bai}~S, {Chen}~F, {Englot}~B. Toward autonomous mapping and exploration for
  mobile robots through deep supervised learning. In: 2017 IEEE/RSJ
  International Conference on Intelligent Robots and Systems (IROS); 2017. p.
  2379--2384.

\bibitem{article23}
{Li}~H, {Zhang}~Q, {Zhao}~D. Deep reinforcement learning-based automatic
  exploration for navigation in unknown environment. IEEE Transactions on
  Neural Networks and Learning Systems. 2020;\hspace{0pt}31(6):2064--2076.

\bibitem{article18}
Takahashi~T, Sun~H, Tian~D, et~al. Learning heuristic functions for mobile
  robot path planning using deep neural networks. In: ICAPS; 2019.

\bibitem{article13}
Qureshi~AH, Simeonov~A, Bency~MJ, et~al. Motion planning networks ; 2018.

\bibitem{article17}
Ariki~Y, Narihira~T. Fully convolutional search heuristic learning for rapid
  path planners ; 2019.

\bibitem{article12}
Bhardwaj~M, Choudhury~S, Scherer~S. Learning heuristic search via imitation ;
  2017.

\bibitem{article19}
Bency~MJ, Qureshi~AH, Yip~MC. Neural path planning: Fixed time, near-optimal
  path generation via oracle imitation. 2019 IEEE/RSJ International Conference
  on Intelligent Robots and Systems (IROS). 2019
  Nov;\hspace{0pt}\urlprefix\url{http://dx.doi.org/10.1109/IROS40897.2019.8968089}.

\bibitem{article25}
DecompositionHowie, ChosetDepartment, Us~A, et~al. Coverage path planning : The
  boustrophedon cellular; 1997.

\bibitem{article1}
Fang~B, Ding~J, Wang~Z. Autonomous robotic exploration based on frontier point
  optimization and multistep path planning. IEEE Access. 2019
  04;\hspace{0pt}PP:1--1.

\bibitem{article8}
{Carlone}~L, {Lyons}~D. Uncertainty-constrained robot exploration: A
  mixed-integer linear programming approach. In: 2014 IEEE International
  Conference on Robotics and Automation (ICRA); 2014. p. 1140--1147.

\bibitem{article9}
Mei~Y, Lu~YH, Lee~C, et~al. Energy-efficient mobile robot exploration. Vol.
  2006; 02; 2006. p. 505 -- 511.

\bibitem{article10}
{Gautam}~A, {Murthy}~JK, {Kumar}~G, et~al. Cluster, allocate, cover: An
  efficient approach for multi-robot coverage. In: 2015 IEEE International
  Conference on Systems, Man, and Cybernetics; 2015. p. 197--203.

\end{thebibliography}

\end{document}